%% file: main.tex
\documentclass[runningheads,a4paper]{llncs}

\usepackage[T1]{fontenc}
\usepackage{lmodern}         % vector Type-1 fonts; avoids Type3/EC bitmaps
\usepackage{graphicx}
\usepackage{booktabs}
\usepackage{tabularx}
\usepackage{amsmath,amssymb}
\usepackage{xspace}
\usepackage[protrusion=true,expansion=false]{microtype}
\usepackage{url}
\usepackage[hidelinks]{hyperref}

\makeatletter
\renewcommand\section{\@startsection{section}{1}{\z@}%
                                  {-10\p@ \@plus -2\p@ \@minus -2\p@}%
                                  {6\p@ \@plus 2\p@ \@minus 2\p@}%
                                  {\normalfont\large\bfseries\boldmath
                                   \rightskip=\z@ \@plus 8em\pretolerance=10000 }}
\renewcommand\subsection{\@startsection{subsection}{2}{\z@}%
                                     {-8\p@ \@plus -2\p@ \@minus -2\p@}%
                                     {4\p@ \@plus 2\p@ \@minus 2\p@}%
                                     {\normalfont\normalsize\bfseries\boldmath
                                      \rightskip=\z@ \@plus 8em\pretolerance=10000 }}
\renewcommand\paragraph{\@startsection{paragraph}{4}{\z@}%
                                    {-6\p@ \@plus -2\p@ \@minus -2\p@}%
                                    {-0.4em \@plus -0.1em \@minus -0.1em}%
                                    {\normalfont\normalsize\itshape}}
\makeatother

\let\oldthebibliography\thebibliography
\renewcommand{\thebibliography}[1]{\oldthebibliography{#1}%
  \setlength{\itemsep}{-2pt plus 0.5pt}%
  \setlength{\parsep}{0pt}}

\renewcommand{\footnotesep}{6pt}

\newcommand{\CoC}{\textsc{CoC}\xspace}
\newcommand{\BIC}{BIC\xspace}

\newif\ifarxiv
\arxivtrue
\newcommand\blfootnote[1]{%
  \begingroup
  \renewcommand\thefootnote{}\footnote{#1}%
  \addtocounter{footnote}{-1}%
  \endgroup
}

\title{When and How Severely: Scenario-Specific Safety
  Envelopes for Driving VLAs}
\titlerunning{Scenario-Specific Safety Envelopes for Driving VLAs}

\author{Abhinaw Priyadershi\inst{1}\orcidID{0009-0007-7120-7972} \and
Jelena Frtunikj\inst{2}\orcidID{0009-0002-5559-1672}}
\authorrunning{A.~Priyadershi and J.~Frtunikj}

\institute{NVIDIA Corporation, Santa Clara, CA, USA\\
\email{apriyadershi@nvidia.com} \and
NVIDIA GmbH, Munich, Germany\\
\email{jfrtunikj@nvidia.com}}

\begin{document}

\maketitle

\ifarxiv
\blfootnote{This version of the contribution has been accepted for publication, after peer review, but is not the Version of Record and does not reflect post-acceptance improvements or any corrections. It will appear in the proceedings of the SAFECOMP 2026 Workshops (WAISE), published by Springer in Lecture Notes in Computer Science; the Version of Record will be available online (DOI to be assigned). Use of this Accepted Version is subject to the publisher's Accepted Manuscript terms of use: \url{https://www.springernature.com/gp/open-research/policies/accepted-manuscript-terms}.}
\fi

\begin{abstract}
Safety certification of Vision-Language-Action (VLA) driving planners
under ISO~21448 (SOTIF) rests on an Operational Design Domain (ODD)
specification that answers two complementary questions: \emph{when}
does the planner start to fail, and \emph{how severely} does it fail
once it does?
We evaluate Alpamayo~R1, a 10B-parameter open-weight driving VLA,
on 15{,}968 (clip, attack) pairs.
We find a \emph{conservative-aggregate gap}: an aggregate
safe threshold of $\sigma \leq 50$ under a 15\% average
displacement error (ADE) budget masks well-sampled scenarios
that tolerate the top of the tested grid ($\sigma = 70$).
A Gaussian Mixture Model (GMM) on the changed-explanation
subset identifies six discrete severity bands
(\BIC-optimal $k{=}6$),
so two perturbation conditions with the same mean error can
differ materially in their share of \emph{high-severity}
(C4/C5) failures.
Joining the two analyses on the same corpus surfaces a finding
neither yields in isolation: the scenarios with the loosest
noise thresholds are not those with the lowest high-severity
rate: STOP\_SIGNAL concentrates roughly $4\times$ the
C4/C5 share of LANE\_KEEPING despite tolerating a larger
$\sigma$. A deployable SOTIF ODD specification for driving
VLAs therefore requires a two-dimensional safety envelope, not
a single aggregate value per hazard.
\end{abstract}

\input{sections/intro}
\input{sections/related_work}
\input{sections/setup}
\input{sections/thresholds}
\input{sections/severity}
\input{sections/synthesis}
\input{sections/discussion}

\bibliographystyle{splncs04}
\bibliography{refs}

\end{document}

%% file: sections/intro.tex
\section{Introduction}
\label{sec:intro}

Consider a driving Vision-Language-Action (VLA) planner certified
as safe up to Gaussian noise $\sigma \leq 50$.
At an intersection, traffic lights and cross-traffic are
high-contrast objects that survive noise well past that level;
the planner stops or proceeds correctly even at $\sigma = 70$.
On the same road, a precision-lateral maneuver (nudging around
a parked vehicle) requires centimeter-level lane-boundary
detection, and the planner may fail at $\sigma = 30$.
A single noise tolerance covers both situations, yet one is
over-protected and the other under-protected.
ISO~21448 (SOTIF)~\cite{ISO21448} formalizes safety-relevant
operating conditions as the Operational Design Domain (ODD):
the set of conditions under which the planner is validated to
operate safely.
Any ODD specification that treats noise tolerance as a single
scalar across all driving scenarios inherits the mismatch above.
VLA-based planners are not yet covered by automotive safety
standards, and the analysis below is exploratory rather than a
compliance procedure; we use ISO~21448 as the closest available
frame of reference for ODD specification.

A second, equally important question arises once a perturbation
does provoke a failure: how bad is the failure?
VLAs produce natural-language Chain-of-Causation (\CoC)
explanations alongside predicted trajectories, e.g.,
\emph{``Slow down because the lead vehicle is braking ahead.''}
Prior work~\cite{PaperI} found that when sensor perturbations
change the \CoC, trajectory error increases $5.3\times$
(point biserial $r_\mathrm{pb} = 0.53$, Cohen's $d = 1.12$,
$n = 15{,}968$).
However, the \emph{magnitude} of text change has near-zero
predictive power (Pearson $r = -0.027$): a complete \CoC
rewrite and a single-word substitution produce statistically
identical trajectory damage.
Binary \CoC change is informative; continuous similarity is not.
This mismatch is also a gap in current robustness
benchmarks~\cite{ROBODRIVE2025,DRIVEBENCH,ROBODRIVEVLM}, which
report aggregate degradation metrics across perturbation
conditions but do not decompose failures by severity.
If trajectory outcomes are discrete, clustered into a small
number of severity bands, such curves mask the actual
failure structure.

A deployable SOTIF ODD specification for a driving VLA
needs to answer both questions jointly.
We evaluate Alpamayo~R1~\cite{ALPAMAYO}, a 10B-parameter
open-weight driving VLA, on 15{,}968 (clip, attack) pairs.
Our contributions are threefold.
(i)~\textbf{When}: per-scenario safe noise thresholds diverge
from the aggregate $\sigma \leq 50$. Four of six
well-sampled categories ($n \geq 30$) reach the maximum tested
$\sigma = 70$, at least $1.4\times$ above the aggregate. Bootstrap
resampling separates strongly-above from moderately-above.
(ii)~\textbf{How severely}: changed-\CoC trajectory errors are
organized into six discrete severity bands
(\BIC-optimal $k{=}6$, stable across 20 restarts).
Two conditions with identical mean error can place different
shares of failures in the high-severity C4 and C5 bands.
(iii)~\textbf{Two-dimensional safety envelope}: joining both
analyses on the same corpus surfaces a finding neither yields
in isolation: the scenarios with the loosest thresholds are
not those with the lowest high-severity rate.
STOP\_SIGNAL and INTERSECTION (both $\sigma \leq 70$)
concentrate 11.8\% and 8.8\% of their changed-\CoC failures
in C4/C5, against 2.9\% for LANE\_KEEPING ($\sigma \leq 50$).

\paragraph{Relation to prior work.}
We separate inheritance, novelty, and methodological
difference relative to~\cite{PaperI}:
\begin{itemize}
  \item \textbf{Inherited:} the 1,996-scenario Alpamayo R1 corpus,
    the aggregate dose-response ($R^2 = 0.957$), and the binary
    \CoC amplification ($5.3\times$ on changed-\CoC pairs).
  \item \textbf{New here:} per-scenario thresholds
    $\sigma^\star(s)$ with bootstrap 95\% CIs
    (Sec.~\ref{sec:thresholds}), a six-band GMM severity
    decomposition (Sec.~\ref{sec:severity}), and a
    two-dimensional ODD envelope pairing $\sigma^\star(s)$ with
    $P(C4\!\cup\!C5\mid\mathrm{coc\_changed})$
    (Sec.~\ref{sec:synthesis}).
  \item \textbf{Methodologically different:}~\cite{PaperI} is
    descriptive and aggregate, treating \CoC change as a single
    binary indicator over the pooled corpus; this paper
    conditions on scenario and severity to produce
    safety-specification primitives.
\end{itemize}
Setup is self-contained in Section~\ref{sec:setup}.

%% file: sections/related_work.tex
\section{Related Work}
\label{sec:related}

\paragraph{VLA Robustness Under Sensor Degradation.}
We identify that prior adversarial robustness research in
autonomous driving has targeted perception components (Eykholt
et~al.~\cite{EYKHOLT2018}; ADvLM~\cite{ADVLM2024}).
The RoboDrive challenge~\cite{ROBODRIVE2025} corrupts
perception inputs for BEV, segmentation, and depth tasks;
RoboDriveVLM~\cite{ROBODRIVEVLM} extends this to VLMs with
11 scenarios (6 sensor, 5 prompt) and 64{,}559 trajectory
cases. DriveBench~\cite{DRIVEBENCH} evaluated 12 VLMs across
17 input conditions and found that models rely on textual
priors over visual grounding under degradation.
The shared limitation is that all three report
\emph{aggregate} performance rather than decomposing results
by driving scenario or trajectory-failure severity.
Our prior work~\cite{PaperI} establishes a
scenario$\times$attack $\Delta$ADE map and the binary \CoC
signal, but reports raw means rather than SOTIF-style
noise-tolerance bounds and does not characterize the shape of
the trajectory-error distribution.

\paragraph{ODD Specification and Discrete Reasoning.}
ISO~21448~\cite{ISO21448} structures the safety argument for
automated driving around the ODD.
Torfah et~al.~\cite{TORFAH22} learn runtime monitors that
predict ODD exits for black-box systems but do not address
whether ODD boundaries should differ by scenario or how
severity should be quantified once a boundary is crossed.
Chain-of-Thought prompting~\cite{COT2022} motivates \CoC
explanations in driving VLAs~\cite{ALPAMAYO}, and Mamou
et~al.~\cite{MAMOU20} show that linguistic categories in deep
language representations emerge as geometrically separable
manifolds; we find a related discrete-band structure in the
trajectory-error outcomes of a driving VLA under sensor
perturbation.
VLAs more broadly extend beyond driving: RT-2~\cite{RT2_2023}
transfers web knowledge to robotic control, and
Senna~\cite{SENNA} is an open-weight candidate for
cross-architecture replication.

\paragraph{Two-Dimensional Safety Envelopes.}
Existing SOTIF-flavored specifications pair an ODD attribute
with a single pass/fail criterion: a scenario is ``inside
the ODD'' if an aggregate metric stays below
threshold~\cite{ROBODRIVE2025,DRIVEBENCH}, or a runtime monitor
declares it so~\cite{TORFAH22}.
Two properties observed here make this treatment insufficient:
the threshold itself is scenario-dependent
(Sec.~\ref{sec:thresholds}), and the trajectory-error
distribution is discrete rather than smooth
(Sec.~\ref{sec:severity}), so two attacks at equal mean error
can place different shares of failures into high-severity bands.
Our two-dimensional envelope (Table~\ref{tab:envelope}) is
orthogonal to monitor learning: it shapes the ODD attribute
such a monitor tracks, rather than replacing the monitor.

%% file: sections/setup.tex
\section{Experimental Setup}
\label{sec:setup}

\subsection{Model, Dataset, and Perturbations}

We evaluate Alpamayo~R1~\cite{ALPAMAYO}, a 10B-parameter
driving VLA with open weights, on 1{,}996 real-world driving
clips from the NVIDIA PhysicalAI-AV
dataset~\cite{PhysicalAI} (publicly available).
Each inference step produces a 64-waypoint trajectory and a
natural-language \CoC{} explanation (cf.\ Section~\ref{sec:intro}).

Eight sensor perturbations span three corruption families:
additive Gaussian noise at $\sigma \in \{10, 30, 50, 70\}$,
photometric scaling (darkening $0.4\times$, brightening
$1.6\times$), and volumetric fog at
$\alpha \in \{0.3, 0.7\}$.
All perturbations are applied synchronously across all camera
views and operate at the pixel level; ego-pose and sensor
metadata are unchanged. Each family affects a different 
stage of the vision pipeline
(sensor readout, exposure, atmospheric transmission).
Each clip is tested under all eight conditions plus one clean
baseline, yielding $1{,}996 \times 8 = 15{,}968$ perturbed
(clip, attack) pairs.
We use \emph{attacked} and \emph{perturbed} (and their noun
forms) interchangeably throughout.

\subsection{Scenario Taxonomy}

We define seven scenario categories by the planning decision
each clip requires (Table~\ref{tab:scenarios}), covering 1{,}558
of the 1{,}996 clips; the remaining 438 belong to an ``OTHER''
category excluded from per-scenario analysis but retained in
full-corpus summaries.
Scenario labels are drawn from the PhysicalAI-AV dataset
metadata~\cite{PhysicalAI}, which annotates each clip with its
primary driving maneuver.
We assign a complexity rank (1 = simplest, 7 = most complex) to
each category based on the number of interacting agents,
decision branches, and required spatial precision in this corpus.

\begin{table}[t]
  \centering\small
  \caption{Scenario distribution, sorted by complexity rank
      (1$=$simplest, 7$=$most complex). NUDGE\_RIGHT is a
      precision-lateral pilot sub-category; TURN\_RIGHT requires
      small-sample caution.}
  \label{tab:scenarios}
  \begin{tabular}{@{}lccc@{}}
    \toprule
    Scenario & $n$ clips & Rank & Clean ADE (m) \\
    \midrule
    LANE\_KEEPING   & 213 & 1 & 1.31 \\
    NUDGE\_RIGHT    &   7 & 2 & 1.94 \\
    FOLLOW\_VEHICLE & 475 & 3 & 1.75 \\
    PASSING         & 177 & 4 & 1.54 \\
    TURN\_RIGHT     &  40 & 5 & 3.11 \\
    STOP\_SIGNAL    & 302 & 6 & 2.47 \\
    INTERSECTION    & 344 & 7 & 2.12 \\
    \bottomrule
  \end{tabular}
\end{table}

\subsection{Metrics}

For each (clip, attack) pair we record two trajectory
quantities and two text quantities.
Average displacement error (ADE, meters) is the mean waypoint
distance between the predicted trajectory and ground truth;
it is computed separately for the clean and perturbed runs.
\emph{ADE degradation} is the difference
$\mathrm{ADE}_\mathrm{attacked} - \mathrm{ADE}_\mathrm{clean}$,
both measured against the same ground-truth trajectory.
Trajectory L2 distance is the aggregated waypoint-level
displacement between the clean and perturbed predicted
trajectories themselves, over the 64-step horizon.
For each pair we also record a binary flag
$\mathrm{coc\_changed} \in \{0,1\}$ and Jaccard token similarity
$J \in [0,1]$ between the clean and perturbed \CoC.
Prior work~\cite{PaperI} found that a binary \CoC change
predicts trajectory deviation while text-change magnitude does
not; we therefore record Jaccard token overlap as the standard
continuous similarity counterpart to the binary flag, enabling
the magnitude-vs-deviation check in Section~\ref{sec:severity}.
Section~\ref{sec:thresholds} uses ADE degradation to define
per-scenario safe noise thresholds; Section~\ref{sec:severity}
uses the trajectory L2 distance to characterize failure-severity
structure among changed-\CoC pairs.

\paragraph{Per-scenario safe threshold.}
We define the per-scenario safe noise threshold - a SOTIF-inspired margin - as the maximum
$\sigma$ at which mean
$\mathrm{ADE}_\mathrm{attacked} < 1.15 \times
\mathrm{ADE}_\mathrm{clean}$, i.e., the attacked trajectory
stays within 15\% of clean performance.
Applying this budget to the aggregate dose-response
from~\cite{PaperI} ($R^2 = 0.957$) yields $\sigma \leq 50$
as the aggregate baseline we compare against.

\paragraph{Binary \CoC signal.}
Prior work~\cite{PaperI} established that a binary \CoC change
predicts trajectory deviation with $r_\mathrm{pb} = 0.53$
(Cohen's $d = 1.12$, $5.3\times$ damage amplification on the
full 15{,}968-pair corpus).
We use the per-scenario \CoC flip rate as a secondary fragility
metric and, in Section~\ref{sec:severity}, restrict the
severity analysis to the 5{,}443 changed-\CoC pairs.

At the aggregate level, ADE degradation increases monotonically
with Gaussian noise severity ($R^2 = 0.957$~\cite{PaperI});
Section~\ref{sec:thresholds} decomposes that aggregate into
per-scenario safe thresholds and shows that they diverge
substantially from the aggregate in both directions.

\paragraph{Reproducibility.}
Alpamayo~R1 has open weights~\cite{ALPAMAYO} and PhysicalAI-AV
is publicly available~\cite{PhysicalAI}; the perturbation grid,
threshold rule, and GMM setup ($k{=}6$ from BIC, full
covariance, 20 random restarts) are documented in
Sections~\ref{sec:setup} through~\ref{sec:severity}.
Per-frame logs and clip-level annotations are internal,
but the qualitative findings (conservative-aggregate gap,
six-band severity structure, threshold-vs-severity
disjunction) should recover on any comparable real-world
driving corpus under the same pipeline.

%% file: sections/thresholds.tex
\section{When: Per-Scenario Noise Thresholds}
\label{sec:thresholds}

\subsection{The Conservative-Aggregate Gap}

ISO~21448~\cite{ISO21448} requires the ODD to state the
conditions under which the planner is validated as safe.
If perturbation tolerance varies across scenarios, a single
aggregate $\sigma$ over-protects some scenarios and
under-protects others; the ODD entry for sensor noise must
therefore be scenario-conditional.

\begin{figure}[t]
  \centering
  \includegraphics[width=0.82\columnwidth]{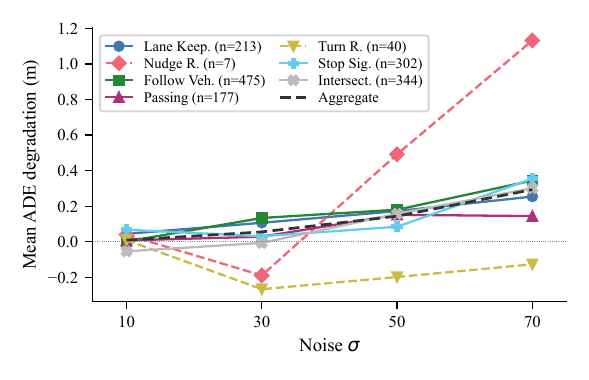}
  \caption{Per-scenario dose-response curves: mean ADE
    degradation versus Gaussian noise $\sigma$ for each
    scenario and the $n$-weighted aggregate.
    Four well-sampled scenarios ($n \geq 30$) stay below the
    15\% ADE budget at $\sigma{=}70$; LANE\_KEEPING and
    FOLLOW\_VEHICLE turn over near the aggregate
    $\sigma{=}50$.
    NUDGE\_RIGHT (pilot, $n{=}7$) spikes at $\sigma{=}70$;
    TURN\_RIGHT runs negative as a small-sample artifact.}
  \label{fig:dose}
\end{figure}

A SOTIF ODD entry for sensor noise, in widespread practice,
assigns one $\sigma$ bound per hazard; the data reject that
practice.
Fig.~\ref{fig:dose} plots mean ADE degradation versus
Gaussian noise $\sigma$ per scenario, and
Table~\ref{tab:thresholds} reports the per-scenario safe
thresholds obtained by applying the 15\% budget to these
curves.
Four of six scenarios with $n \geq 30$ clips (PASSING,
TURN\_RIGHT, STOP\_SIGNAL, INTERSECTION) stay safe up to the
tested maximum $\sigma \leq 70$, while LANE\_KEEPING and
FOLLOW\_VEHICLE match the aggregate $\sigma \leq 50$
from~\cite{PaperI}. These thresholds are right-censored at
the grid ceiling; the true scenario-specific margins for the
four looser scenarios may be higher than $\sigma = 70$.

Deploying at $\sigma \leq 50$ uniformly therefore wastes
operational capacity in four of six well-sampled scenarios
that could safely handle at least 40\% more noise; we call
this heterogeneity the \emph{conservative-aggregate gap}.
Bootstrap resampling ($10{,}000$ clip-level resamples) adds
graduated confidence: TURN\_RIGHT (97.5\% of resamples at
$\sigma{=}70$) and PASSING (86.6\%) are strongly above the
aggregate; STOP\_SIGNAL (56.2\%) and INTERSECTION (58.0\%)
are suggestive but not definitive.
Recomputed on the 1{,}558 classified clips only, the
aggregate remains $\sigma \leq 50$ (mean ADE at $\sigma{=}70$:
$2.214$~m $>$ budget $2.209$~m).
NUDGE\_RIGHT ($n{=}7$ clips, 56~pairs) breaks in the
opposite direction ($\sigma \leq 30$); the bootstrap CI is
uninformative and we treat it as a pilot finding requiring
$n \geq 30$ replication.
The gap is stable under budget choice: at 10\% and 20\% it
persists, with each scenario's threshold shifting by at most
one grid step and NUDGE\_RIGHT tightest at every budget.

\begin{table}[t]
  \centering\small
  \caption{Per-scenario ODD thresholds (15\% ADE budget) with
    bootstrap 95\% CI ($10{,}000$ resamples). Point estimates
    for 4/6 well-sampled scenarios exceed the aggregate
    (Clause~5.2 ODD specification).}
  \label{tab:thresholds}
  \begin{tabular}{@{}lcccc@{}}
    \toprule
    Scenario & $n$ & Max $\sigma$ & 95\% CI & vs.\ agg. \\
    \midrule
    NUDGE\_RIGHT  &   7 & $\leq 30$\textsuperscript{*} & $[30,70]$ & Tighter \\
    LANE\_KEEPING & 213 & $\leq 50$ & $[10,70]$ & Matches \\
    FOLLOW\_VEH.  & 475 & $\leq 50$ & $[30,70]$ & Matches \\
    PASSING       & 177 & $\leq 70$ & $[30,70]$ & Looser \\
    TURN\_RIGHT   &  40 & $\leq 70$ & $[50,70]$ & Looser \\
    STOP\_SIGNAL  & 302 & $\leq 70$ & $[50,70]$ & Looser \\
    INTERSECTION  & 344 & $\leq 70$ & $[50,70]$ & Looser \\
    \bottomrule
    \multicolumn{5}{@{}l@{}}{\scriptsize\textsuperscript{*}Pilot ($n{=}7$).}
  \end{tabular}
\end{table}

\subsection{Precision Demand vs.\ Decision Complexity}

The natural hypothesis that complex scenes are more fragile
under noise is not supported here.
The more noise-tolerant scenarios are the spatially coarse ones
(INTERSECTION, STOP\_SIGNAL): their relevant visual features
(traffic lights, cross-traffic presence, stop-sign state) are
high-contrast objects that survive pixel-level noise.
Simpler scenarios such as FOLLOW\_VEHICLE require continuous
distance estimation, which is more sensitive to noise but
still tolerates $\sigma \leq 50$.
NUDGE\_RIGHT is the counter-example: despite a low complexity
rank, it demands centimeter-level lateral precision for
lane-boundary detection.
The distinction is between \emph{precision demand} and
decision complexity, and it is the key practical insight:
safety engineers cannot use scene complexity as a proxy for
risk; they must measure precision demand per scenario.
We report the overall complexity-vs.-threshold correlation
only as a hypothesis for larger-scale replication.

\subsection{Scenario-Dependent Attack Profiles}

The scenario$\times$attack $\Delta\mathrm{ADE}$ interactions in~\cite{PaperI} compound this: fog hits signal-compliance
scenarios (STOP\_SIGNAL, PASSING) hardest, while heavy
Gaussian noise concentrates on NUDGE\_RIGHT
($\Delta\mathrm{ADE}\!=\!{+}1.13$~m at $\sigma{=}70$).
Accordingly, a deployment-grade SOTIF ODD table may require
scenario$\times$attack entries rather than the single
per-scenario threshold in Table~\ref{tab:thresholds}.
The \emph{when} of VLA failure is therefore
scenario-specific; whether scenario-specificity also tracks
the \emph{severity} of failures is the question of
Section~\ref{sec:severity}, and their joint implication is
taken up in Section~\ref{sec:synthesis}.

%% file: sections/severity.tex
\section{How Severely: Discrete Failure Bands}
\label{sec:severity}

Section~\ref{sec:thresholds} showed \emph{when} the VLA
starts to fail by scenario.
We now examine \emph{how severely} it fails once it does.
Of the 15{,}968 (clip, attack) pairs, 10{,}525 (66\%) preserve
the \CoC text (mean L2 $\approx 4.1$~m); the remaining 5{,}443
(34\%) change it (mean L2 $= 21.8$~m, a $5.3\times$ increase).
The point-biserial correlation between the binary
$\mathrm{coc\_changed}$ flag and L2 is $r_\mathrm{pb} = 0.53$
(Cohen's $d = 1.12$, $p \approx 0$), and at the perturbation
level the \CoC change rate and mean L2 deviation correlate at
$r = 0.994$~\cite{PaperI}.

\subsection{Text-Change Magnitude Is Not Predictive}

The magnitude of the text change carries little information
about error severity. Among the 5{,}443 changed pairs,
Jaccard token similarity and L2 displacement correlate at
$r = -0.027$ (Spearman $\rho = 0.010$): a complete \CoC
rewrite ($J < 0.10$, $n{=}556$) and a mid-range edit
($0.10 \leq J \leq 0.90$, $n{=}4{,}731$) produce
indistinguishable trajectory damage (MWU $p = 0.257$, Cohen's
$d = -0.026$; post-hoc power $>99\%$ to detect $d \geq 0.10$).
The natural-language explanation tells you \emph{that} the
model switched, not \emph{how far} it deviated. Binary
\CoC-change detection is therefore both simpler and more
effective than magnitude-based metrics for trajectory-damage
prediction in this setting.
Jaccard measures token overlap; an embedding-based similarity
(e.g., sentence-transformer cosine) could distinguish paraphrases
from semantically distinct rewrites inside the mid-Jaccard band.
We did not test that and leave the comparison to future work.

\subsection{GMM Severity Bands}

\begin{figure}[t]
  \centering
  \includegraphics[width=\textwidth]{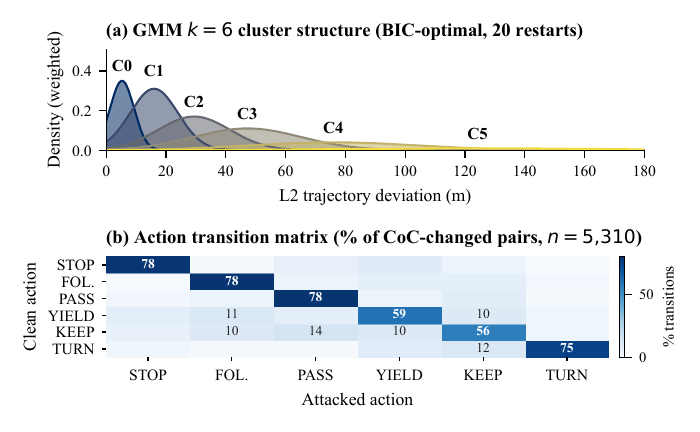}
  \caption{The \emph{Discrete Reasoning Menu}.
    \textbf{(a)} GMM clusters on L2 of changed pairs
    ($n{=}5{,}443$); the sharp \BIC kink at $k{=}5 \to 6$
    is consistent with the discrete structure.
    \textbf{(b)} action transition matrix ($n{=}5{,}310$
    non-OTHER pairs): 56 to 78\% of \CoC changes per action
    are surface rewrites (diagonal dominance).}
  \label{fig:menu}
\end{figure}

A Gaussian Mixture Model~\cite{GMM} fitted to the L2
distribution of the 5{,}443 changed-\CoC pairs yields
\BIC-optimal $k{=}6$ components, selected in all 20 random
restarts.
The \BIC improvement from adding the 6th component is
$2\times$ larger than from adding a 7th or 8th, indicating
that the 6th captures genuine structure while further
components do not.
Five-fold cross-validation over $k{=}1,\ldots,8$ likewise
favors $k{=}6$: held-out log-likelihood for $k{=}6$ exceeds
$k{=}5$ in all five folds.
Fits use scikit-learn's \texttt{GaussianMixture}
(full covariance, \texttt{max\_iter=200}; 20 restarts iterate
seeds 0 to 19 at \texttt{n\_init=3}).
Table~\ref{tab:clusters} lists the six components, with
means from the final GMM fit on the combined corpus;
Fig.~\ref{fig:menu}(a) visualizes them.

\begin{table}[t]
  \centering\small
  \caption{GMM $k{=}6$ cluster structure on changed pairs
    ($n{=}5{,}443$); component means fixed globally.
    Supplies the Clause~6.4 severity attribute
    $P(C4\!\cup\!C5\mid\mathrm{coc\_changed})$.}
  \label{tab:clusters}
  \begin{tabular}{@{}lccl@{}}
    \toprule
    Band & Mean L2 & Share & Interpretation \\
    \midrule
    C0 &   5.3~m & 35\% & Near-stop; minor wobble \\
    C1 &  16.0~m & 31\% & Creep; noticeable drift \\
    C2 &  29.4~m & 17\% & Follow-adjustment \\
    C3 &  47.2~m & 11\% & Moderate maneuver \\
    C4 &  76.0~m &  4\% & Hard maneuver \\
    C5 & 124.2~m &  1\% & Emergency \\
    \bottomrule
  \end{tabular}
\end{table}

As a concrete illustration, noise\_70 and bright produce
similar mean L2 errors among changed pairs (23.3~m vs.\
21.0~m), but noise\_70 places 7.5\% of failures in the
high-L2 C4/C5 bands versus 4.9\% for bright, a 53\% gap.
We label C4/C5 \emph{high-severity} rather than
\emph{catastrophic}: in the absence of collision,
rule-violation, or near-miss ground truth, the label is a
proxy for downstream severity, not a claim about realized
outcomes.

Per-scenario GMMs ($k{=}1 \ldots 6$) select $k{=}2$ to $k{=}5$
within each scenario; no individual scenario is unimodal.
Cram\'er's $V = 0.135$ ($\chi^2(45) = 498.7$, $p < 0.001$)
indicates a statistically significant but weak
scenario-dependence in cluster assignment: the global
$k{=}6$ structure does not merely reflect six scenario types
each contributing one characteristic error value.
We treat the $k{=}6$ structure as an observable property of
the L2 distribution, not as a claim about internal VLA
decision mechanisms.

\subsection{Convergent Evidence}

Four complementary analyses support the discrete
interpretation.
The most informative is the \emph{noisy proxy}: mid-range
Jaccard values ($0.10$ to $0.90$, 86.9\% of changed pairs)
produce statistically identical damage as complete rewrites
($d = -0.026$, $p = 0.257$).
Mid-range texts are therefore paraphrases of the same discrete
decision, not intermediate-severity failures.
Cross-attack stability ($\rho = 0.347$, ICC $= 0.35$~\cite{ICC})
indicates that fragility is a clip-level
property: 26.5\% of clips never flip under any of the eight
attacks while 2.5\% always flip.
The bimodality coefficient
BC $= 0.642 > 0.555$~\cite{BIMODALITY} independently supports
a non-continuous distribution.
The transition matrix in Fig.~\ref{fig:menu}(b)
(restricted to the 5{,}310 non-OTHER changed-\CoC pairs
that carry a classified scenario label; the GMM above uses
all 5{,}443 including OTHER clips) shows
that 71\% of \CoC changes preserve the original action
category, so the band a failure lands in is not determined
by the action word alone.
The trajectory-outcome space is therefore better described
as a discrete partition rather than a continuous manifold, and two
perturbation conditions with identical mean error can carry
different high-severity exposure.
Section~\ref{sec:synthesis} intersects this severity
structure with the per-scenario thresholds of
Section~\ref{sec:thresholds}.

%% file: sections/synthesis.tex
\section{A Two-Dimensional Safety Envelope}
\label{sec:synthesis}

We combine the two axes from
Sections~\ref{sec:thresholds} and~\ref{sec:severity}, which answer
two questions on the same corpus: \emph{when} the 15\% ADE
budget is first exceeded, and \emph{how severely} failures
distribute once a perturbation flips the \CoC.
Neither axis alone suffices: a threshold without a severity
profile is silent on whether violations are high-severity, and
a severity profile without a threshold is silent on how often
they occur.

\subsection{Joint Distribution: Scenario $\times$ Severity}

\begin{figure}[t]
  \centering
  \includegraphics[width=\textwidth]{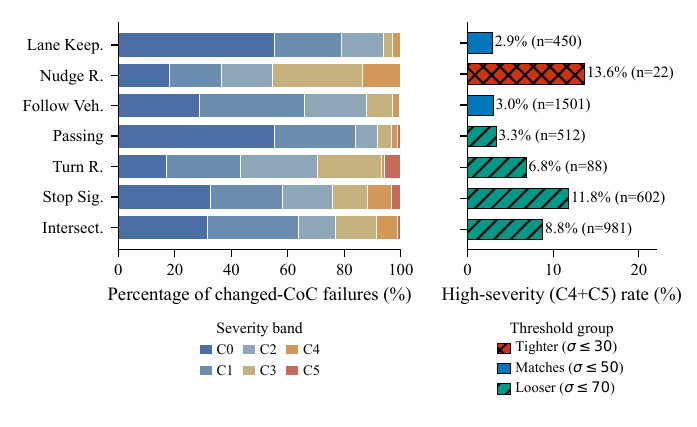}
  \caption{Joint scenario $\times$ severity distribution on
    changed-\CoC pairs ($n{=}4{,}156$ in seven classified
    scenarios).
    \textbf{Left:} percentage of each scenario's failures in
    bands C0 through C5.
    \textbf{Right:} high-severity (C4/C5) rate per scenario;
    bar color encodes the threshold group from
    Table~\ref{tab:thresholds} (red: tighter; blue: matches;
    green: looser than aggregate).
    STOP\_SIGNAL and INTERSECTION (both $\sigma \leq 70$)
    concentrate 11.8\% and 8.8\% of failures in C4/C5,
    against 2.9\% for LANE\_KEEPING ($\sigma \leq 50$).}
  \label{fig:joint}
\end{figure}

Restricting to the 4{,}156 changed-\CoC pairs in the seven
classified scenarios yields the joint distribution in
Fig.~\ref{fig:joint}.

\paragraph{The two axes are not redundant.}
Fig.~\ref{fig:joint} (right) and Table~\ref{tab:thresholds}
are not the same ranking.
STOP\_SIGNAL and INTERSECTION sit at the top of both: a loose
threshold ($\sigma \leq 70$) \emph{and} the highest
high-severity rates (11.8\% and 8.8\%) among well-sampled
scenarios.
LANE\_KEEPING is the reverse: a tighter threshold
($\sigma \leq 50$) but only 2.9\% high-severity share when
failures do occur.
A single-axis ODD specification would collapse these scenarios
onto the same point and obscure a safety-relevant asymmetry.

\paragraph{Threshold robustness implies frequency, not magnitude.}
Consistent with the precision-demand pattern in
Section~\ref{sec:thresholds}, INTERSECTION and STOP\_SIGNAL
are spatially coarse decisions whose high-contrast cues
survive pixel-level noise, so failures are rare. But when one
occurs, a stop/proceed flip maps to a large trajectory
difference. Lane-keeping and vehicle-following degrade earlier
on subtle cues, but produce small continuous deviations rather
than categorical flips.
NUDGE\_RIGHT ($n{=}7$ clips, 22 changed pairs) remains the
precision outlier, tightest threshold and highest C4/C5
share (13.6\%); the small sample makes both numbers pilot
estimates.

\subsection{ODD Specification as a Two-Dimensional Entry}

We propose a two-dimensional ODD entry
(Table~\ref{tab:envelope}): the threshold $\sigma^\star$ at which the 15\% ADE
budget is first exceeded, and the high-severity rate
$P(C4\!\cup\!C5 \mid \mathrm{coc\_changed})$ from the joint
distribution in Fig.~\ref{fig:joint}.

\begin{table}[t]
  \centering\small
  \caption{Two-dimensional ODD entry per scenario
    (Clause~6.5 acceptance artifact: both columns must meet
    the residual-risk budget). $\sigma^\star$: safe threshold at 15\% ADE;
    $P(C4\!\cup\!C5 \mid \mathrm{coc\_changed})$: fraction of
    changed-\CoC pairs in the two highest bands.}
  \label{tab:envelope}
  \begin{tabular}{@{}lcccc@{}}
    \toprule
    Scenario & $n_\mathrm{chg}$ & $\sigma^\star$ &
      $P(C4\!\cup\!C5)$ & Quadrant \\
    \midrule
    NUDGE\_RIGHT  &   22 & $\leq 30$\textsuperscript{*} & 13.6\% & Tight $+$ severe\\
    LANE\_KEEPING &  450 & $\leq 50$ &  2.9\% & Match $+$ mild \\
    FOLLOW\_VEH.  & 1501 & $\leq 50$ &  3.0\% & Match $+$ mild \\
    PASSING       &  512 & $\leq 70$ &  3.3\% & Loose $+$ mild \\
    TURN\_RIGHT   &   88 & $\leq 70$ &  6.8\% & Loose $+$ moderate \\
    STOP\_SIGNAL  &  602 & $\leq 70$ & 11.8\% & Loose $+$ severe \\
    INTERSECTION  &  981 & $\leq 70$ &  8.8\% & Loose $+$ severe \\
    \bottomrule
    \multicolumn{5}{@{}l@{}}{\scriptsize\textsuperscript{*}Pilot ($n{=}7$ clips).}
  \end{tabular}
\end{table}

Table~\ref{tab:envelope} differs from Table~\ref{tab:thresholds}
in a safety-relevant way.
An operator reading
Table~\ref{tab:thresholds} alone treats STOP\_SIGNAL and
PASSING symmetrically (both $\sigma \leq 70$), whereas
Table~\ref{tab:envelope} places them in different quadrants
(PASSING ``loose $+$ mild,'' STOP\_SIGNAL ``loose $+$ severe''),
warranting different graduated-response policies despite
sharing a threshold.
Within a pair, text-change magnitude does not predict severity
(Sec.~\ref{sec:severity}); across scenarios, the threshold
does not predict severity either; no scalar summary fully
specifies a scenario's ODD entry.

Concretely, the two-dimensional entry targets five
ISO~21448:2022 clauses~\cite{ISO21448}. Per-scenario
$\sigma^\star(s)$ with bootstrap 95\% CIs
(Table~\ref{tab:thresholds}) populates Clause~5.2 as an ODD
attribute of the specification, and its onset doubles as a
Clause~7.3 triggering condition for the same scenario. The
high-severity rate $P(C4\!\cup\!C5\mid\mathrm{coc\_changed})$
from the six-band GMM (Table~\ref{tab:clusters}) supplies the
Clause~6.4 severity attribute, and the pair
$(\sigma^\star,\pi^\star)$ in Table~\ref{tab:envelope} forms
the Clause~6.5 residual-risk acceptance criterion.
Table~\ref{tab:assurance} packages these as a Clause~12
release argument.

%% file: sections/discussion.tex
\section{Discussion and Future Work}
\label{sec:discussion}

\subsection{Implications for SOTIF ODD Specification}

We argue that the aggregate $\sigma \leq 50$ being overly
conservative for four of six scenarios has a direct consequence
for ISO~21448 compliance~\cite{ISO21448}: a single SOTIF hazard
entry for ``sensor noise'' leaves operational capacity on
the table.
The discrete six-band structure is complementary: two
perturbation types with identical mean trajectory error can
carry different risk when one concentrates failures in C0/C1
and the other in C4/C5.
Table~\ref{tab:assurance} maps the two into a single
goal/strategy/evidence argument (Evidence nodes E1--E3,
Assumptions A1--A2).

\begin{table}[t]
  \centering\scriptsize
  \caption{ISO~21448 assurance mapping for the two-dimensional
    SOTIF ODD entry. $\sigma^\star(s)$: per-scenario threshold;
    $\pi^\star$: high-severity-rate budget.}
  \label{tab:assurance}
  \begin{tabularx}{\columnwidth}{@{}lX@{}}
    \toprule
    \textbf{Goal}
      & For each scenario $s$, the planner's behavior under
        sensor noise is within the SOTIF ODD.\\
    \textbf{Strategy}
      & Argue over a two-dimensional envelope: a
        triggering-condition bound (Clause~7.3) paired with
        a residual-risk acceptance criterion (Clause~6.5).\\
    \textbf{E1}
      & Per-scenario $\sigma^\star(s)$ with bootstrap 95\% CI
        (Table~\ref{tab:thresholds}; Clause~5.2 ODD
        specification).\\
    \textbf{E2}
      & Clause~6.4 severity attribute
        $P(C4\!\cup\!C5\mid\mathrm{coc\_changed})$ from the
        six-band GMM (Table~\ref{tab:clusters}).\\
    \textbf{E3}
      & Clause~6.5 acceptance: $s$ passes iff
        $\sigma \leq \sigma^\star(s)$ and
        $P(C4\!\cup\!C5\mid s) \leq \pi^\star$
        (Table~\ref{tab:envelope}); release argument per
        Clause~12.\\
    \textbf{A1, A2}
      & 15\% ADE budget is stable under $\pm$5\% sensitivity
        (Section~\ref{sec:thresholds}); C4/C5 is a proxy for
        downstream severity pending collision or
        rule-violation tie-in. Selecting $\pi^\star$ from a
        target societal risk level is out of scope.\\
    \bottomrule
  \end{tabularx}
\end{table}

The procedure is post hoc and requires no additional
training: compute $\sigma^\star(s)$ under the chosen ADE
budget and report the per-scenario C4/C5 rate from a GMM
on the changed-\CoC subset. Two conditions with similar mean
error but different C4/C5 rates receive different safety
ratings.

\subsection{Limitations and Future Work}

All analyses use a single VLA (Alpamayo~R1) on a single
dataset (1{,}996 clips) under eight pixel-level perturbations;
temporal corruptions (frame drops, sensor latency) and
adversarial patches are untested.
The 15\% ADE degradation budget is reasonable but arbitrary;
operational deployment requires domain-expert SOTIF hazard
review.
NUDGE\_RIGHT ($n{=}7$) is a pilot with an uninformative
bootstrap CI and needs $n \geq 30$ replication.
We treat the six-component GMM as an observational summary of
the L2 distribution on this corpus, not a claim about latent
internal decision states or a general property of driving VLAs.
A second open-weight VLA and a second dataset are needed to
separate model- from corpus-specific structure.
Alpamayo~R1.5 offers an immediate same-family cross-version
test, and Senna~\cite{SENNA} is a cross-architecture
replication candidate.
Extending the perturbation battery to temporal corruptions,
together with severity-graded runtime monitoring that gates a
C0 to C5 cluster posterior on the per-scenario $\sigma^\star$,
are the immediate follow-ups.

Driving VLAs in zero-shot use do not expose calibrated
internal uncertainty estimates, and standard DNN uncertainty
quantification (ensembles, MC dropout, evidential heads)
requires weight access or retraining that an integrator of a
third-party model does not have. A black-box dose-response
with severity-band characterization is therefore the safety
envelope available to a Tier~1 or OEM under those constraints.

The ``two is better than one'' finding generalizes as follows:
an aggregate safety metric is lossy whenever the failure space
has dimensions that do not move together. Threshold and
severity decorrelate across scenarios in this corpus, so
neither projects onto the other; testing this pattern on
additional VLAs and perturbation classes is the natural
follow-up.

\section{Conclusion}\label{sec:conclusion}

We show that one noise threshold does not fit all driving
scenarios, and that one mean-error curve does not describe how
severely failures bite.
A deployable SOTIF ODD specification for a driving VLA
therefore requires a two-dimensional entry per scenario: a
noise threshold (\emph{when} the envelope is violated) paired
with a severity distribution (\emph{how far} it extends).
On this corpus, the loosest thresholds do not match the
lowest high-severity rates.

\begin{credits}
\subsubsection{\discintname}
All authors are NVIDIA employees; no external funding was received.
\end{credits}